# Face Recognition Using Discrete Cosine Transform for Global and Local Features

Aman R. Chadha, Pallavi P. Vaidya, M. Mani Roja
Department of Electronics & Telecommunication
Thadomal Shahani Engineering College
Mumbai, INDIA
aman.x64@gmail.com

*Abstract*— Face Recognition using Discrete Cosine Transform (DCT) for Local and Global Features involves recognizing the corresponding face image from the database. The face image obtained from the user is cropped such that only the frontal face image is extracted, eliminating the background. The image is restricted to a size of 128 × 128 pixels. All images in the database are gray level images. DCT is applied to the entire image. This gives DCT coefficients, which are global features. Local features such as eyes, nose and mouth are also extracted and DCT is applied to these features. Depending upon the recognition rate obtained for each feature, they are given weightage and then combined. Both local and global features are used for comparison. By comparing the ranks for global and local features, the false acceptance rate for DCT can be minimized.

I. INTRODUCTION

A face recognition system is essentially an application [1] intended to identify or verify a person either from a digital image or a video frame obtained from a video source. Although other reliable methods of biometric personal identification exist, for e.g., fingerprint analysis or iris scans, these methods inherently rely on the cooperation of the participants, whereas a personal identification system based on analysis of frontal or profile images of the face is often effective without the participant's cooperation or intervention. One of the many ways for automatic identification or verification is by comparing selected facial features from the image and a facial database. This technique is typically used in security systems. For e.g., the technology could be used as a security measure at ATM's; instead of using a bank card or personal identification number, the ATM would capture an image of the person's face, and compare it to his/her photo in the bank database to confirm the identity of the relevant person. On similar lines, this concept could also be extrapolated to computers; by using a web cam to capture a digital image of a person, the face could replace the commonly used password as a means to log-in and thus, authenticate oneself. Given a large database of images and a photograph, the problem is to select from the database a small set of records such that one of the image records matched the photograph. The success of the method could be measured in terms of the ratio of the answer list to the number of records in the database. The recognition problem is made difficult by the great variability in head rotation and tilt, lighting intensity and angle, facial expression, aging, etc. Some other attempts at facial recognition by machines have allowed for little or no variability in these quantities. A general statement of the problem of machine recognition of faces can [2] be formulated as: given a still or video image of a scene, identify or verify one or more persons in the scene using a stored database of faces. Available collateral information such as race, age, gender, facial expression, or speech may be used in narrowing the search. The solution to the problem involves segmentation of faces, feature extraction from face regions, recognition, or verification. In identification problems, the input to the system is an unknown face, and the system reports back the determined identity from a database of known individuals, whereas in verification problems, the system needs to confirm or reject the claimed identity of the input face.

Some of the various applications of face recognition include driving licenses, immigration, national ID, passport, voter registration, security application, medical records, personal device logon, desktop logon, human-robot-interaction, human-computer-interaction, smart cards etc. Over the last 3 decades, many methods of face recognition have been proposed. Face recognition is such a challenging yet interesting problem that it has attracted researchers who have different backgrounds: pattern recognition, neural networks, computer vision, and computer graphics, hence the literature is vast and diverse. Often, a single system involves techniques motivated by different principles. The usage of a mixture of techniques makes it difficult to classify these systems based on what types of techniques they use for feature representation or classification. To have clear categorization, we follow the holistic and local features approach [2]. Specifically, we have following categories:

*1) Holistic matching methods:* These methods use the whole face region as a raw input to the recognition system. One of the most widely used representations of the face region is Eigenpictures, which is inherently based on principal component analysis.

*2) Feature-based matching methods:* Generally, in these methods, local features such as the eyes, nose and mouth are first extracted and their locations and local statistics are fed as inputs into a classifier.

*3) Hybrid methods:* It uses both local features and whole face region to recognize a face. This method could potentially offer the better of the two types of methods.

Our aim is to extract local features from a given frontal face. The local features are left eye, right eye, nose and mouth. These local features will be extracted manually. Discrete Cosine Transform (DCT) will be applied to each of these local features individually and also to the global features. Finally, the results obtained in both cases will be compared.

Four popular face recognition methods, namely Principle Component Analysis (PCA), Spectroface, Independent Component Analysis (ICA), and Gabor jet are selected and three popular face databases, namely Yale database, Olivetti Research Laboratory (ORL) database and FERRET database, are selected for evaluation in the paper [3]. It proposes to make use of both local features and global features for face recognition and performs experiments in combining two global feature face recognition algorithms, namely, PCA, Spectroface, and two local feature algorithms, namely, Gabor wavelet and ICAs. The experimental results show that the 'rank 1' accuracy ranges from 79.5% to 85.5% and the 'rank 2' accuracy of overall performance for the proposed idea is 92.5%. A face recognition system in which interesting feature points in face are located by Gabor filters is discussed in [4]. Then the filtered image is multiplied with a 2D Gaussian to focus on the center of the face and avoid extracting features at face contour. This Gabor filtered and Gaussian weighted image is then searched for peaks, which are called feature points used for recognition. The accuracy is 83.4% taking all feature points which is better than that obtained in case of single feature strategy.

## II. IDEA OF THE PROPOSED SOLUTION

For creating the database standard, BioID database [5] is used. Frontal images are extracted from the database. Frontal face image extraction is carried out in order to reduce the effect of varying backgrounds on the proposed face recognition system. The frontal face image is acquired. From the given frontal image, local features like eyes, nose and mouth region are extracted manually. DCT is applied to each of these features. The DCT coefficients of these regions are stored. These coefficients are then used for comparison. The recognition rates obtained with these local features are compared to the recognition rate obtained when DCT is applied to the global image.

### A. Discrete Cosine Transform

DCT is an accurate and robust face recognition system and using certain normalization techniques, its robustness to variations in facial geometry and illumination can be increased [6],[7]. An alternative holistic approach to face recognition is discrete cosine transform. Face normalization techniques were also incorporated in the face recognition system discussed. Namely, an affine transformation was used to correct scale, position, and orientation changes in faces. It was seen that tremendous improvements in recognition rates could be achieved with such normalization. Illumination normalization was also investigated extensively. Various approaches to the problem of compensating for illumination variations among faces were designed and tested, and it was concluded that the recognition rate of the specific system was sensitive to many of these approaches. This was because the faces in the databases used for the tests were uniformly illuminated and these databases contained a wide variety of skin tones. That is, certain illumination normalization techniques had a tendency to make all faces have the same overall grey-scale intensity, and they thus resulted in the loss of much of the information about the individuals' skin tones. A complexity comparison between the DCT and the Karhunen-Loeve transform (KLT) is of interest [1]. In the proposed method, training essentially means computing the DCT coefficients of all the database faces. On the other hand, using the KLT, training entails computing the basis vectors of the transformation. This means that the KLT is more computationally expensive with respect to training. However, once the KLT basis vectors have been obtained, it may be argued that computing the KLT coefficients for recognition is trivial. But this is also true of the DCT, with the additional provision that the DCT may take advantage of very efficient computational algorithms.

DCT is a well-known signal analysis tool used in compression due to its compact representation power [8]. It is known that the KLT is the optimal transform in terms of information packing, however, its data dependent nature makes it infeasible to implement in some practical tasks. Moreover, DCT closely approximates the compact representation ability of the KLT, which makes it a very useful tool for signal representation both in terms of information packing and in terms of computational complexity due to its data independent nature. DCT helps separate the image into parts (or spectral sub-bands) of differing importance (with respect to the image's visual quality). DCT is conceptually similar to Discrete Fourier Transform (DFT), in the way that it transforms a signal or an image from the spatial domain to the frequency domain.

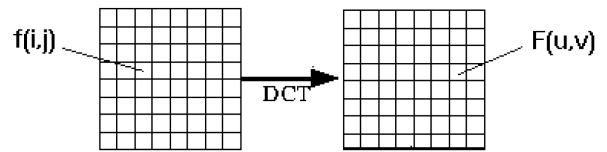

Figure 1. Image transformation from spatial domain to frequency domain

### B. Discrete Cosine Transform Encoding

The general equation for a 1D (N data items) DCT is defined as follows:

$$F(u) = \sqrt{\frac{2}{N}} \sum_{i=0}^{N-1} A(i) * \cos\left(\frac{u(2i+1)\pi}{2N}\right) * f(i) \quad (1)$$

where,

$$A(i) = \begin{cases} \frac{1}{\sqrt{2}} & \text{for } u = 0 \\ 1 & \text{otherwise} \end{cases}$$

f(i) is the input sequence.

The general equation for a 2D (N × M image) DCT is defined as follows:

$$F(u,v) = \sqrt{\frac{2}{N}}\sqrt{\frac{2}{M}}\sum_{i=0}^{N-1} A(i) * \cos\left(\frac{u(2i+1)\pi}{2N}\right) * \sum_{j=0}^{M-1} A(j) * \cos\left(\frac{v(2j+1)\pi}{2M}\right) * f(i,j) \quad (2)$$

where,

$$A(i) = \begin{cases} \frac{1}{\sqrt{2}} & \text{for } u = 0 \\ 1 & \text{otherwise} \end{cases}$$

$$A(j) = \begin{cases} \frac{1}{\sqrt{2}} & \text{for } v = 0 \\ 1 & \text{otherwise} \end{cases}$$

f(i, j) is the 2D input sequence.

The basic encoding operation of the DCT is as follows:

- The size of the input image is N × M.
- f(i, j) is the intensity of the pixel at x(i, j).
- F(u, v) is the DCT coefficient for the pixel at x(i, j).
- For most images, much of the signal energy lies at low frequencies. These appear in the upper left corner of the DCT.
- Compression is achieved since the lower right values represent higher frequencies, and are often small enough to be neglected with little visible distortion. The output array of DCT coefficients contains integers; these can range from -1024 to 1023.

Computationally, it is easier to implement and also efficient to consider the DCT as a set of basis functions which given a known input array size, for e.g., 8 × 8, can be pre-computed and stored. This involves simply computing values for a convolution mask (8 × 8 window) that get applied. The DCT coefficients are calculated using (2).

### III. IMPLEMENTATION STEPS

The database consists of a set of images of 25 people. There are four test images and a single training or registered image. There are 4 different test-methods: GLOBAL, LOCAL, GLOBAL+LOCAL and GLOBAL AND LOCAL. The images are converted to average intensity with respect to the registered image stored in the database, i.e., the images are normalized.

*A. Normalization*

Since the facial images are captured at different instants of the day or on different days, the intensity for each image may exhibit variations. To avoid these light intensity variations, the test images are normalized so as to have an average intensity value with respect to the registered image. The average intensity value of the registered images is calculated as summation of all pixel values divided by the total number of pixels. Similarly, average intensity value of the test image is calculated. The normalization value is calculated as:

$$\text{Normalization Value} = \frac{\text{Average value of registered image}}{\text{Average value of test image}} \quad (3)$$

This value is multiplied with each pixel of the test image. Thus we get a normalized image having an average intensity with respect to that of the registered image. The entire image is of size 128 × 128 pixels. Upon applying DCT and performing zigzag scanning, we obtain 16384 coefficients, of which 64 coefficients are taken into account while matching.

*B. Zigzag Scanning*

The purpose of Zigzag Scanning is to:

- Map 128 × 128 to a 1 × 16384 vector.
- Group low frequency coefficients present at the top of the vector.

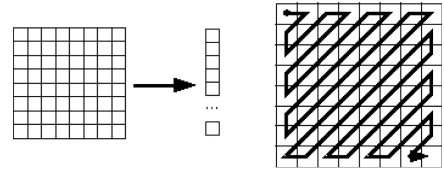

Figure 2. Zigzag scanning

The eye regions are cropped with a size of 16 × 16 pixels. Next, the nose and mouth region are cropped. The mouth region contains the outline of the lips. This region is rectangular in shape. The nose region is 25 × 40 pixels and the mouth region is 30 × 50 pixels. Local features such as eyes, nose and mouth are extracted manually from the given face image. The image is used as an input. The centre eye pixels are located and a region of 16 × 16 pixels is extracted which covers the eye region. The nose region and the mouth region are extracted as shown in Fig. 4. The maximum margin for the nose region is 40 × 25 pixels while that for the mouth region is 50 × 30 pixels. Only the lower portion of nose region, i.e., the region around the nostrils, should be extracted. Similarly, for the mouth region, only the outline of the lips should be considered. Local feature extraction, especially eye extraction, helps reduce the effect of varying characteristics such as pose, expressions etc. on the face recognition system.

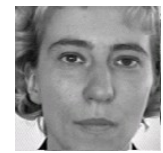

Figure 3. Test image

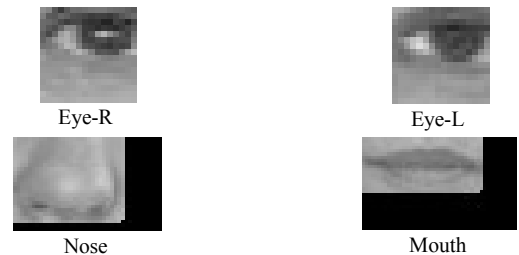

Eye-R    Eye-L

Nose    Mouth

Figure 4. Local feature extraction

After extracting the features, the images are compared using all the methods mentioned above. The comparison is done by taking the Euclidean distance between the test and registered image. For e.g., we take 50 coefficients of test image and 50 coefficients of registered image. The Euclidean distance of each of the 50 coefficients of the test image and 50 coefficients of the registered image are calculated. Add all the Euclidean distance of the 50 coefficients. Let the value of this addition be 'X'. Since there are 15 registered images, each of the test images will be compared with the 15 registered images. Thus we have $X_1, X_2, X_3, \ldots, X_{15}$ values. All these 15 values of 'X' are sorted in ascending order, and the one with the minimum value of 'X' is given as the 'rank 1'. The next in the order is given 'rank 2', 'rank 3', till 'rank 15'. This 'rank 1' image is regarded as the best match. If the 'rank 1' image is the same as the input image, the person has been recognized correctly. Thus, the recognition rate is calculated as the ratio of number of images correctly recognized to the total number of images tested. The number of coefficients is varied and the recognition rate is calculated for each of them using the following equation:

$$\text{Recognition rate} = \frac{\text{Number of correctly recognized persons}}{\text{Total number of persons tested}} \quad (4)$$

IV. RESULTS

TABLE I. DCT RESULTS

| Features | Recognition Rate |
|---|---|
| Entire Image | 88.25% |
| Eye-R | 87.18% |
| Eye-L | 86.1% |
| Nose | 56.2% |
| Mouth | 52.35% |

*A. Global Features*

Recognition rate using global features, i.e., taking the entire image, has been tabulated as follows:

TABLE II. RECOGNITION RATE USING GLOBAL FEATURES

| Global Features | Recognition Rate | |
| | *Without normalized image* | *With normalized image* |
|---|---|---|
| 64 coeffecients | 88.25% | 92.5% |

*B. Local Features*

Recognition rate using only local features, i.e., taking eyes, nose and mouth templates, has been tabulated as follows:

TABLE III. RECOGNITION RATE USING LOCAL FEATURES

| Local Features | Recognition Rate | |
| | *Without normalized image* | *With normalized image* |
|---|---|---|
| DCT | 87.18% | 90.2% |

*C. Using AND Logic*

The ranks of both the global feature and local features are compared. If both the ranks are '1' only then is the person accepted, else the person's entry is termed as 'invalid'. Thus the false acceptance rate is zero in this case. The corresponding recognition rate has been tabulated as follows:

TABLE IV. RECOGNITION RATE USING AND LOGIC

| AND Logic | Recognition Rate | |
| | *Without normalized image* | *With normalized image* |
|---|---|---|
| DCT | 80.52% | 82.35% |

*D. Combining Local and Global Features*

The local templates and global templates are combined together, with different weights assigned to each template. The corresponding recognition rate has been tabulated as follows:

TABLE V. RECOGNITION RATE ON COMBINING LOCAL AND GLOBAL FEATURES

| Combination | Recognition Rate | |
| | *Without normalized image* | *With normalized image* |
|---|---|---|
| DCT | 90.4% | 94.5% |

V. CONCLUSION

When using local features for recognition, the false acceptance rate should be minimized and false rejection rate should be maximized as compared to that of global features. The recognition rate for local features and the recognition rate for global features using DCT is calculated. Comparison between DCT global features and DCT local features is done. The recognition rate improves when images are normalized. When local and global features are combined, DCT gives a relatively high recognition rate.

REFERENCES

[1] Ziad M. Hafed, "Face Recognition Using DCT", International Journal of Computer Vision, 2001, pp. 167-188.
[2] W. Zhao, R. Chellappa, "Face Recognition: A Literature Survey", ACM Computing Surveys, Vol. 35, No. 4, December 2003, pp. 399-458, pp. 9-11.
[3] J. Huang, P. Yuen, J. Lai, "Face Recognition Using Local and Global Features", EURASIP Journal on Applied Signal Processing 2004:4, pp. 530-541.
[4] E. Hjelmas, "Feature-Based Face Recognition", Department of Informatics, University of Oslo, pp. 1-5.
[5] BioID Database, www.bioid.com, October 2007.
[6] M. Tistarelli, L. Akarun, "Report on face state of art", BioSecure, Biometrics for secure Authentication, April 2005, pp. 1-76.
[7] M. Ahmad, T. Natarajan and K. R. Rao, "Discrete Cosine Transform", IEEE Trans. Computers, 1974, pp. 90-94.
[8] H. Ekenel, R. Stiefelhagen, "Block Selection in the Local Appearance-based Face Recognition Scheme", Interactive Systems Labs, Computer Science Department, University Karlsruhe, Germany, pp. 1-6.